\def\year{2022}\relax
\documentclass[letterpaper]{article} 
\usepackage{aaai22}  
\usepackage{tabu}
\usepackage{times}  
\usepackage{helvet}  
\usepackage{courier}  
\usepackage[hyphens]{url}  
\usepackage{graphicx} 
\usepackage{amssymb}
\usepackage{natbib}  
\usepackage{caption} 
\DeclareCaptionStyle{ruled}{labelfont=normalfont,labelsep=colon,strut=off} 
\usepackage{algorithm}
\usepackage{algorithmic}

\newcommand{\figref}[1]{Fig.~\ref{#1}}
\newcommand{\tabref}[1]{Tab~\ref{#1}}
\newcommand{\eref}[1]{Eq.~(\ref{#1})}

\newcommand{\figureref}[1]{Figure~\ref{#1}}
\newcommand{\tableref}[1]{Table~\ref{#1}}

\usepackage{color}

\usepackage{newfloat}
\usepackage{listings}
\floatstyle{ruled}
\newfloat{listing}{tb}{lst}{}
\floatname{listing}{Listing}

\title{Intermediate and Future Frame Prediction of Geostationary Satellite Imagery With Warp and Refine Network}
\author {
    Minseok Seo\textsuperscript{\rm 1},
    Yeji Choi \textsuperscript{\rm 1}\thanks{corresponding author},
    Hyungon Ryu \textsuperscript{\rm 2},
    Heesun Park \textsuperscript{\rm 3}
    Hyungkun Bae \textsuperscript{\rm 1},
    Hyesook Lee \textsuperscript{\rm 3},
    Wanseok Seo \textsuperscript{\rm 2},
}
\affiliations {
    \textsuperscript{\rm 1} SI Analytics, 
    \textsuperscript{\rm 2} NVIDIA,
    \textsuperscript{\rm 3} National Institute of Meteorological Sciences (NIMS) \\
    \textsuperscript{*}yejichoi@si-analytics.ai
}

\usepackage{bibentry}

\begin{document}

\maketitle

\begin{abstract}
Geostationary satellite imagery has applications in climate and weather forecasting, planning natural energy resources, and predicting extreme weather events.
For precise and accurate prediction, higher spatial and temporal resolution of geostationary satellite imagery is important.
Although recent geostationary satellite resolution has improved, the long-term analysis of climate applications is limited to using multiple satellites from the past to the present due to the different resolutions.
To solve this problem, we proposed warp and refine network (WR-Net). WR-Net is divided into an optical flow warp component and a warp image refinement component.
We used the TV-L1 algorithm instead of deep learning-based approaches to extract the optical flow warp component. The deep-learning-based model is trained on the human-centric view of the RGB channel and does not work on geostationary satellites, which is gray-scale one-channel imagery.
The refinement network refines the warped image through a multi-temporal fusion layer.
We evaluated WR-Net by interpolation of temporal resolution at 4 min intervals to 2 min intervals in large-scale GK2A geostationary meteorological satellite imagery. 
Furthermore, we applied WR-Net to the future frame prediction task and showed that the explicit use of optical flow can help future frame prediction.
\end{abstract}

\section{Introduction}
Weather prediction using geostationary satellite imagery indirectly contributes to the earth’s environment in various fields such as planning natural energy resources~\cite{pathak2022fourcastnet} and extreme weather event prediction.
Physically-based numerical weather prediction models have traditionally been used to analyze and forecast weather and climate, but recent developments~\cite{seo2022simple, seo2022domain} in deep learning-based models have made data-driven methods possible to use for weather prediction. 
In the era of the climate crisis, accurate weather prediction is essential to prepare the sudden disasters and to maximize the use of renewable energy such as wind and solar power. Therefore, the spatial and temporal resolution of forecasting results is important to appropriate countermeasures.
Recently, a geostationary-satellite resolution has been improved with 1-2 minute intervals with a pixel size of 1-2 km. However, to integrate multiple satellites from the past to the present for long-term analysis for climate applications, there is a limit to high-resolution analysis because most of the past geostationary satellites have a lower resolution of 10 to 15 minutes.

To solve this problem, Vandal and Nemani~\cite{vandal2021temporal} proposed a temporal interpolation of geostationary satellite imagery with optical flow approach based on Super SloMo~\cite{jiang2018super}.
In the Super SloMo based method, when images $I_{t}$ and $I_{t+1}$ are given, the interpolation frame $I_{i}$ is predicted through optical flow estimation and warp between them.
This method achieved state-of-the-art, but does not extract the optical flow properly when a sample that has not been learned in the training phase is input and is vulnerable to intensity changes such as brightness changes.
To solve this problem, video frame interpolation methods using pre-trained weights for PWC-Net~\cite{sun2018pwc} and FlowNet~\cite{dosovitskiy2015flownet} have been proposed.
However, PWC-Net and FlowNet trained on RGB channels of human-centric view do not work in geostationary satellite imagery.

In this paper, we propose Warp and Refine Network (WR-Net).
WR-Net is divided into optical flow estimation and warp component and refinement component.
In the optical flow estimation and warp component, since deep learning-based supervised optical flow estimation methods cannot be used, we estimate and warp the optical flow through the total variation-L1 algorithm, a non-parametric approach.
After that, the refinement network refines the intensity change through a multi-temporal fusion of the warp frame with $I_{t}$ and $I_{t+1}$ to refine the intensity change, such as the brightness change.
We verified the WR-Net in large-scale GK2A geostationary meteorological satellite imagery, and as a result achieved state-of-the-art.
Furthermore, we extend WR-Net to future frame prediction.
Our motivation is that optical flow includes information such as movement direction and speed, and this will be effective not only for video frame interpolation but also for future frame prediction.
We predicted the frame after 90 min through WR-Net and evaluated it qualitatively.
As a result of qualitative analysis, it was shown that the explicit use of optical flow is helpful for future frame prediction.
\begin{figure*}[t!]
    \centering
    \includegraphics[width=1.4\columnwidth]{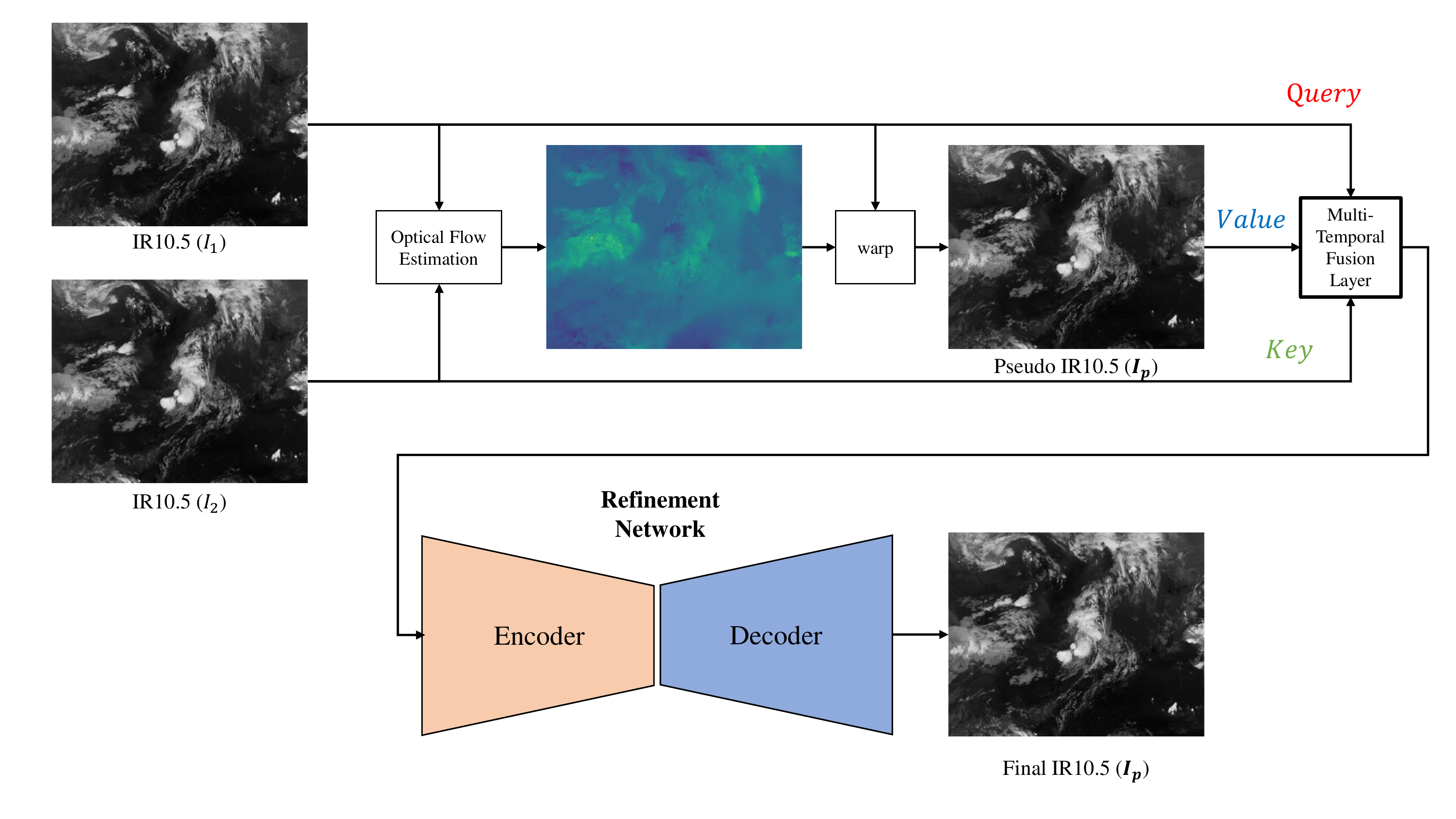}
    \caption{An overview of our proposed WR-Net. WR-Net consists of an optical flow warp component and a refinement component. The optical flow is extracted through the TV-L1 algorithm. The image warp with the extracted optical flow is refined through a multi-temporal fusion network and a refinement network.}
    \label{fig:overview}
\end{figure*}
Our contributions can be summarized as follows: 
\vspace{-2mm}
\begin{itemize}
\setlength\itemsep{0em}
    \item We propose WR-Net to extract optical flow using TV-L1 algorithm and warp and refine it.
    \item Our WR-Net achieves state-of-the-art in geostationary satellite imagery video frame interpolation.
    \item To the best of our knowledge, for the first time, optical flow was explicitly applied to deep learning-based weather forecasting.

\end{itemize}
\section{Method}
In this section, we divide our proposed Warp-and-Refine Networks (WR-Net) into three components and describe them in detail.
The first component, optical flow warp, estimates the optical flow between the $I_{t}$ image and the $I_{t+1}$ image, and warps the $I_{t}$ or $I_{t+1}$ image with the estimated optical flow to predict intermediate or future frames.
Therefore, we propose and use a multi-temporal fusion layer to correct $I_{inter}, I_{future}$ by referring to the intensity values of highly correlated parts in $I_{t}$ and $I_{t+1}$ images. The multi-temporal fusion feature is input to the U-Net based Refine Network to refine the intensity value.

\subsection{Optical Flow Warp}
\begin{figure}[h!]
    \centering
    \includegraphics[width=\columnwidth]{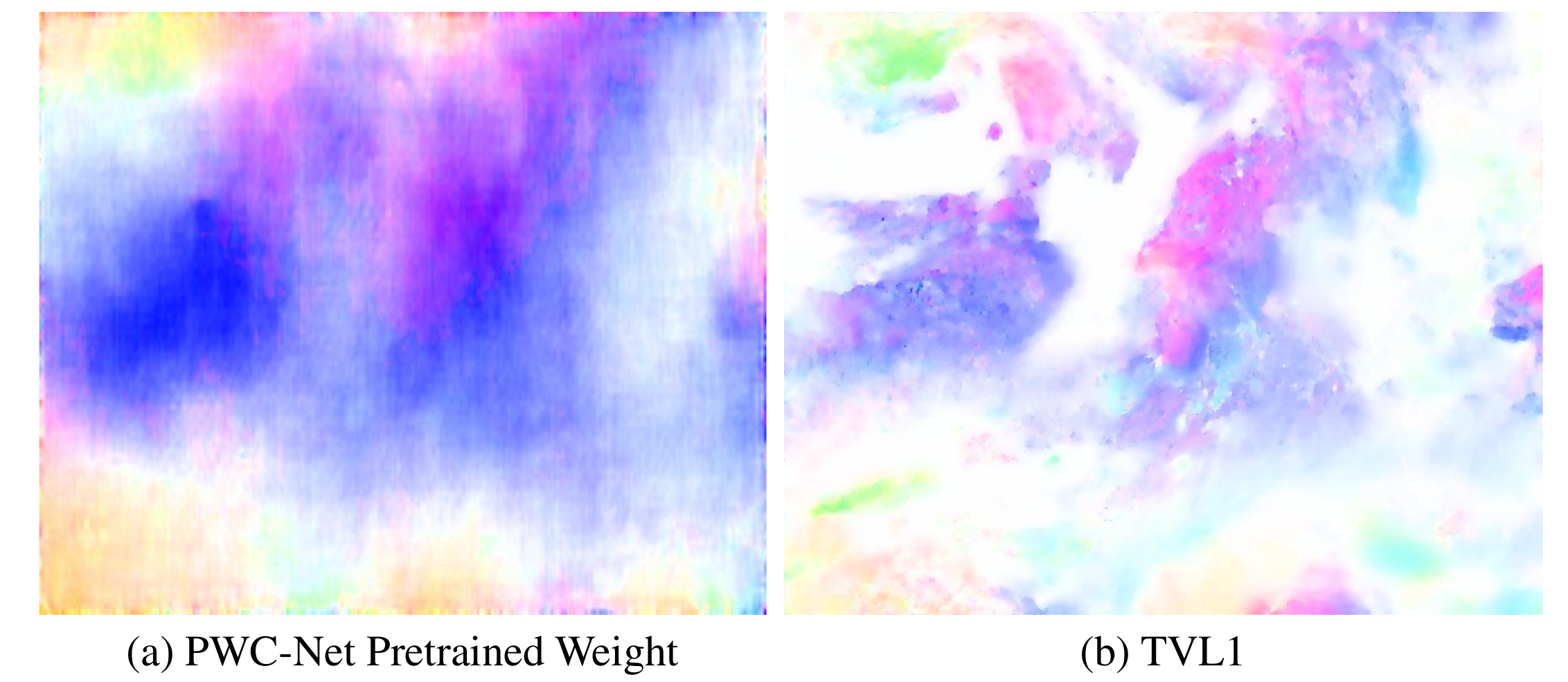}
    \caption{Qualitative comparison of optical flow extracted with PWC-Net and TV-L1 algorithms.}
    \label{fig:optical}
\end{figure}

The optical flow warp method is a recent learning approach in the field of video frame interpolation.
However, as shown in ~\figref{fig:optical}-(a), the existing optical flow estimation network is designed and trained on the human-centric view of RGB images, so the performance of geostationary satellite imagery is greatly reduced.
Therefore, we estimate the optical flow by using the GPU accelerated total variation-L1 (TVL1)~\cite{wedel2009improved} algorithm as shown in ~\figref{fig:optical}-(b) instead of the deep learning-based optical flow estimate network.

When there is $I_{t}$ frame and $I_{t+1}$, the intermediate frame $I_{i}$ and future frame $I_{f}$ can be synthesized through ~\eref{eq:optical_warp_inter} and ~\eref{eq:optical_warp_outer}, respectively.

\begin{equation}
  I_{i} = g(I_{t}, Op(I_{t+1}, I_{t}) \times \alpha),
  \label{eq:optical_warp_inter}
\end{equation}

\begin{equation}
  I_{f} = g(I_{t+1}, Op(I_{t+1}, I_{t}) \times \alpha),
  \label{eq:optical_warp_outer}
\end{equation}
where $Op(.,.)$ is a TV-L1 optical flow estimation algorithm.
The $g(.)$ is a backward warping function.
Parameter $\alpha$ is the interval between time $t$ and $t+1$.
For example, if $\alpha$ is set 0.5, an intermediate frame $I_{t+0.5}$ can be generated, or a future frame $I_{t+1.5}$ can be generated.

\subsection{Multi Temporal Fusion Network}
The warp based on optical flow is not an optimal value because $I_{i}$ or $I{f}$ are generated by linear interpolation of $I_{t}$ or $I{t+1}$.
Therefore, we need to refine the pixel intensity values that cannot be expressed by linear interpolation.
Therefore, we refine the intensity value of $I_{i}$ or $I_{f}$ using the information of $I_{t}$ and $I_{t+1}$.

We propose a multi-temporal fusion layer based on Non-local Neural Networks to converge multi-temporal information.
\begin{figure}[t]
    \centering
    \includegraphics[width=\columnwidth]{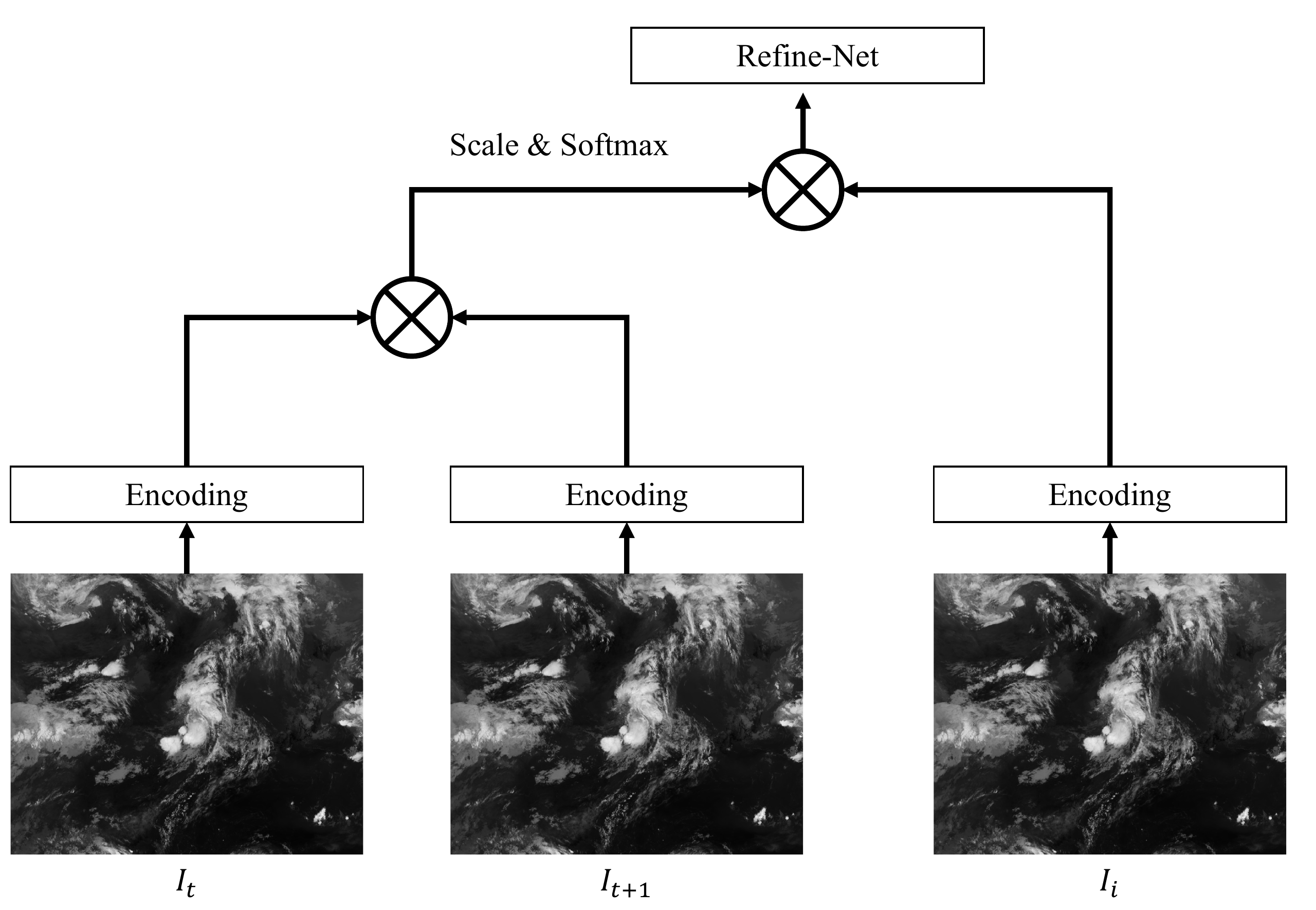}
    \caption{Overview of multi-temporal fusion layers. In video frame interpolation, $I_{t}$ and $I_{t+1}$ are input as query and key. In future frame prediction, replace $I_{i}$ with the predicted future frame.}
    \label{fig:fusion}
\end{figure}
~\figureref{fig:fusion} shows the structure of a multi-temporal fusion layer. As shown in the ~\figref{fig:fusion}, the features with high correlation between $I_{t}$ and $I_{t+1}$ are weighted and multiplied by $I_{i}$. Through this process, $I_{i}$ can be modeled for correlation between $I_{t}$ and $I_{t+1}$.

\subsection{Refinement Network}
Multi-temporal fusion features are input to U-Net~\cite{ronneberger2015u} based Refinement Network. Refinement Network restores the input feature to the ground truth $I_{i_{gt}}$ or $I_{f_{gt}}$.
When the input images $I_{t}$ and $I_{t+1}$ are given, the intermediate frame prediction's objective function of Warp and Refine Network is as follows:
\begin{equation}
  l_{r} = ||I_{i_{gt}}-Refine(MTF(I_{t+1}, I_{t}, I_{i}))||_{1}, 
  \label{eq:vfi_obj}
\end{equation}
where $MTF(. , . , .)$ is a Multi temporal fusion network. The $Refine(.)$ is a refinement network.
For future frame prediction, simply change $I_{i}$ to $I_{f}$ in ~\eref{eq:vfi_obj}.

\section{Experiments}
\paragraph{Dataset} We trained and evaluated WR-Net on a large-scale GK2A geostationary satellite weather observation imagery to evaluate the intermediate and future frame prediction performance.
The GK2A (GEO-KOMPSAT-2A) satellite dataset~\cite{chung2020meteorological} from August 2020 to July 2021 is used for training and validation. The dataset is in 2-minute intervals over the East-Asia regions with 2 km spatial resolution .
Since the GK2A dataset is very large-scale, only 1, 6, 11, 16, 21, and 26 days were selected and used.

We strictly followed the GK2A official user manual for pre-processing the physical value data of GK2A.
The image size of the GK2A dataset is 1950 $\times$ 1550, and in our experiments, only channels of infrared ray (IR) 10.5 $\mu m$, short wave infrared ray (SW) 0.38 $\mu m$, and water vapor (WV) 0.69 $\mu m$ were used.
\paragraph{Training}
We set the time step parameter $\tau$ to 0.25, the attachment weight parameter $\lambda$ to 0.15, and the tightness parameter $\theta$ to 0.3 when using the optical flow with the TV-L1 algorithm.
As for augmentation, we used random crop augmentation with a size of 975 $\times$ 775 and random rotate 90$^{\circ}$ augmentation.
Each input channel of the multi-modal fusion network is embedded with 512 channels, and the U-Net based on VGG16 is used as the refine network. 

All experiments were performed on NVIDIA A100 GPU$\times$8, batch size was 64, adaptive moment estimation (Adam) was used as the optimizer, and the learning rate was set to 1e-4. 
Note that the hyperparameters of the TV-L1 algorithm has a significantly effect on the performance, so it must be adjusted according to the dataset.

\begin{table} 
\centering
\begin{tabu}{c|c|c|c}
\tabucline[1pt]{-}
\textbf{Method}& \textbf{Channels}&\textbf{PSNR $\uparrow$} &\textbf{SSIM $\uparrow$}  \\ \hline
Linear & IR  & 38.667   & 0.745 \\
SSM-T  & IR & 43.285    & 0.831 \\ 
WR-Net (Warp Only)  & IR & 44.213 & 0.904  \\ 
WR-Net (Full)  & IR & \textbf{46.527} & \textbf{0.934}  \\ \hline
Linear & SW  & 38.381   & 0.719 \\
SSM-T  & SW & 45.935    & 0.827 \\ 
WR-Net (Warp Only)  & SW & 48.237 & 0.922  \\
WR-Net (Full)  & SW & \textbf{50.538} & \textbf{0.936}  \\\hline
Linear & WV  & 43.526  & 0.766 \\
SSM-T  & WV & 51.842    & 0.895 \\ 
WR-Net (Warp Only)  & WV & 56.191 & 0.929 \\
WR-Net (full) & WV  & \textbf{58.350}   & \textbf{0.973} \\ 
\tabucline[1pt]{-}
\end{tabu}
\caption{Quantitative experimental results of video frame interpolation of WR-Net on the GK2A geostationary satellite weather observation dataset.}
\label{table:vif}
\end{table}
\paragraph{Video frame interpolation results}
The scale of weather tracking and modeling, ecosystem monitoring and climate change tracking depend on the spatial and temporal resolutions of observations.

However, when we use multiple satellites, the temporal resolution needs to be set to the past geostationary (GEO) satellite resolution at intervals of 10-15 min, although we have a higher resolution dataset with 1-2 min intervals.
To solve this problem, we use WR-Net to interpolate the temporal resolution of geostationary (GEO) satellites and evaluate their performance through peak signal-to-noise ratio (PSNR) and structural similarity index map (SSIM).

We interpolated satellite images at 4min intervals in 2min increments and compared their performance with SSM-T (state-of-the-art method~\cite{vandal2021temporal}) and linear interpolation.

~\tableref{table:vif} shows the results of video frame interpolation experiments. As shown in ~\tabref{table:vif}, WR-Net showed the highest performance in IR 10.5 $\mu m$, SW 0.38 $\mu m$, and WV 0.69 $\mu m$ channels.
In addition, the performance of video frame interpolation by extracting the optical flow through the TV-L1 algorithm and simply warping the optical flow was higher than that of SSM-T.
These experimental results indicate that the video frame interpolation method of superslomo series, which extracts and wraps the optical flow as an unsupervised learning method, is inefficient in the weather observation geostationary satellite.

\begin{figure*}[t!]
    \centering
    \includegraphics[width=1.6\columnwidth]{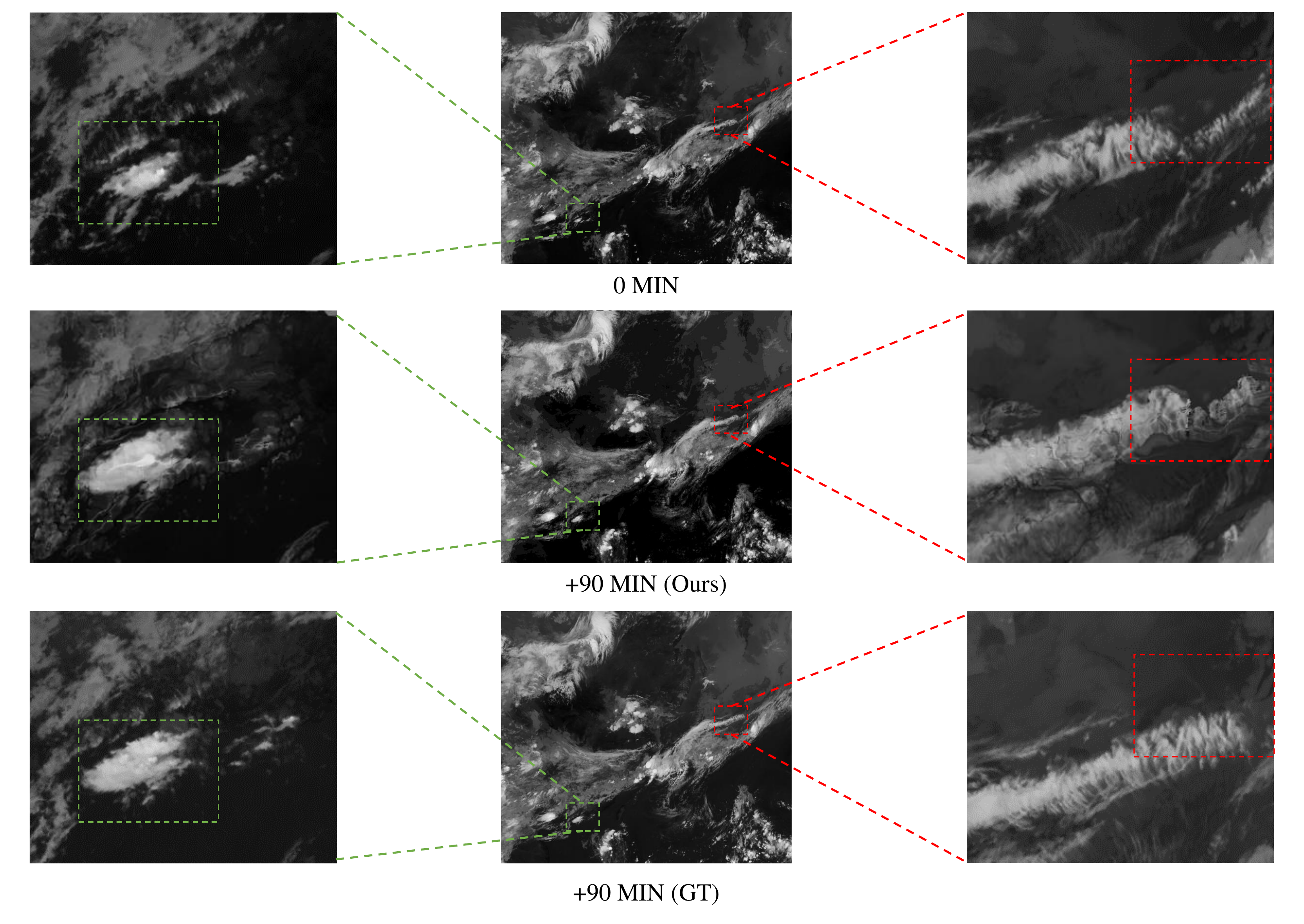}
    \caption{Qualitative result of future frame prediction of WR-Net. WR-Net receives 0min and +10min as input and predicts +20min, and inputs between +10min and predicted +20min again into WR-Net to predict +30min. Repeat this process to predict up to +90min.}
    \label{fig:vis}
\end{figure*}

Finally, the performance of using the refinement network was higher than the optical flow warping of WR-Net alone. These experimental results indicate that the refinement network operated significantly.

\paragraph{Future frame prediction results}
Future frame prediction is a very important field that can be used in various fields such as nowcasting, tracking temperature change, disaster forecasting, and climate change forecasting.
According to this importance, deepmind's deep generative model of rain (DGMR)~\cite{ravuri2021skilful} or FourCastNet has been proposed. All of these methods modeled multi-temporal information, but they were limited to either concatenating and inputting multi-temporal frames to the deep learning model or using LSTM.

Optical flow is a modality that includes movement direction and velocity between two input frames. Especially in the short term, the direction of movement is rarely reversed. Because of these characteristics, movement direction and speed are very important analysis factors in weather forecasting.

Therefore, we applied WR-Net to verify the usability of optical flow in the field of future frame prediction. ~\figureref{fig:vis} is the qualitative result of predicting the frame after 90 minutes through WR-net.

As shown in the green box in ~\figref{fig:vis}, the cloud expansion direction and size were predicted similarly to the actual ground truth.
%
%
These quantitative experimental results indicate that optical flow is very helpful in predicting cloud expansion and direction, and we confirm the possibility of predicting clouds generations with the refinement network (the green box in Fig. 4). However, in some cases, it has limitations to generate newly developed clouds (the red box in Fig. 4).  
\section{Limitation and Future Work}
Our study performed video frame interpolation or future frame prediction using only single channels with short timestamps.
However, tasks such as weather forecasting and climate forecasting can be performed with high performance by fusion of information from various channels as well as a single channel.
In our future work, we plan to study a method using multi-channel information. Also, we will verify WR-Net with tropical cyclone cases to check the generation and dissipation of cloud in the future study.

\section{Conclusion}
We solved the problem that PWC-Net or FlowNet did not work in weather observation satellites through TV-L1 algorithm and proposed WR-Net to achieve state-of-the-art in the field of video frame interpolation.
Also, as far as we know, it showed the possibility of using optical flow for the first time in the field of future frame prediction.
Although our work has only been evaluated by weather observation satellites, we believe that our method has applications in a variety of fields, including climate change and disaster prediction.
We hope that our research will have widespread applications in earth science as well as weather observations.

\newpage
\bibliography{aaai}

\bigskip

\end{document}